\documentclass[11pt]{article}
\usepackage[final]{acl}

\usepackage{times}
\usepackage{latexsym}
\usepackage[T1]{fontenc}
\usepackage[utf8]{inputenc}
\usepackage{microtype}
\usepackage{inconsolata}
\usepackage{graphicx}
\usepackage{amsmath}
\usepackage{amssymb}

\usepackage{booktabs}

\usepackage{multirow} 
\usepackage{xcolor}
\usepackage[normalem]{ulem}
\usepackage[inline]{enumitem}

\usepackage{siunitx}

\newcommand{\mf}[1]{\textcolor{black}{#1}}

\newcommand{\wl}[1]{#1}

\title{Data-Driven Persona-Conditioned Agents for A/B Test Simulation}
\author{Ziyad Benomar\thanks{\ \ Equal contribution.} \\
  Amazon \\ Luxembourg, Luxembourg \\
  \texttt{zbenomar@amazon.lu} \\\And
  Weronika Łajewska\footnotemark[\value{footnote}] \\
  Amazon \\ Luxembourg, Luxembourg\\
  \texttt{lajewska@amazon.lu} \\\AND
  Leonardo Perelli \\
  Amazon \\ Luxembourg, Luxembourg\\
  \texttt{lperelli@amazon.lu} \\\And 
  Saab Mansour \\
  Amazon \\ Barcelona, Spain\\
  \texttt{saabm@amazon.es} \\}

\begin{document}
\maketitle

\begin{abstract}
A/B testing is the gold standard for evaluating product changes, but each experiment requires real user traffic, engineering effort, and weeks of measurement. We propose a simulation framework that predicts A/B test outcomes using LLM-powered agents conditioned on \emph{data-driven} personas grounded in real user behavioral signals. Unlike prior work that relies on synthetic or rule-based personas, our agents are constructed from anonymized behavioral data---activity patterns, engagement signals, and inferred demographics---enabling more faithful population modeling. We frame A/B test simulation as a structured question task and systematically study (i)~question design formats, (ii)~the impact of persona data source and domain alignment, (iii)~the trade-off between per-persona behavioral depth and population diversity, and (iv)~efficient population subsampling. On a benchmark of 40 A/B tests spanning two metric types, our best configuration achieves 0.75--0.80 directional accuracy depending on the test metric, demonstrating that data-driven personas are a viable path toward fast, low-cost experiment pre-screening.
\end{abstract}

\section{Introduction}
\label{sec:intro}

LLMs can emulate human decision-making across tasks from survey response prediction~\citep{Argyle:2023:PA, Aher:2023:ICML} to preference elicitation and behavioral role-play~\citep{Wang:2025:TOIS, Mansour:2025:REALM_paars}. When conditioned on detailed user profiles, LLM-powered agents approximate the judgments of specific population segments, enabling scalable, low-cost simulation of collective user behavior~\citep{Park:2024:arXiv, Bui:2025:ACL}. A particularly compelling application is the simulation of online controlled experiments (A/B tests): if persona-conditioned agents can reliably predict whether users prefer a treatment variant over a control, teams could pre-screen design candidates offline---reducing the time, traffic, and experimentation cost~\citep{Rieder:2026:arXiv_SimAB, Castelo:2026:arXiv_SimGym}.

Online controlled experiments remain the gold standard for validating product changes, yet each test requires sufficient user traffic, engineering effort, and typically weeks of data collection to reach statistical significance~\citep{Kohavi:2009:DMKD}. These costs limit how many ideas teams can evaluate.
Recent work explores LLM-based simulation as a complement to live experimentation.
SimAB~\citep{Rieder:2026:arXiv_SimAB} shows that persona-conditioned agents can predict directional outcomes of A/B tests, but relies on synthetic personas generated from variant screenshots without grounding in real user behavior. SimGym~\citep{Castelo:2026:arXiv_SimGym} replays real traffic traces through browser agents to simulate e-commerce experiments, but requires access to detailed session logs and does not study persona construction. Both systems highlight the promise of simulation, yet leave open how to construct user representations that capture the behavioral diversity of real population.

We present a simulation framework that predicts A/B test outcomes using LLM agents conditioned on \emph{data-driven} personas grounded in real behavioral data. Unlike synthetic personas derived from screenshots or rule-based descriptions~\citep{Rieder:2026:arXiv_SimAB}, ours are constructed from anonymized behavioral signals spanning months of user activity~\citep{Mansour:2025:REALM_paars}. This grounding lets us address questions prior work cannot: how does behavioral depth affect simulation fidelity? What is the trade-off between per-persona richness and population representativeness? How can a persona pool be subsampled without degrading quality? We frame A/B test simulation as a structured question task and study the interplay between question design, persona construction, and population sampling---providing the first empirical analysis of these factors in an A/B experimentation setting.

We evaluate our framework on a benchmark of 40 A/B tests spanning two metric types---click-through rate (CTR) and subscriptions---and organize experiments around four research questions:
\begin{enumerate}[leftmargin=1em, itemsep=0pt, parsep=0pt, topsep=3pt]
    \item \emph{How does the question format used to elicit agent preferences affect simulation accuracy?} 
    \item \emph{Do data-driven personas outperform synthetic personas for A/B test simulation?} 
    \item \emph{What is the trade-off between per-persona behavioral depth and population-level diversity?} 
    \item \emph{How aggressively can the persona pool be subsampled while preserving simulation quality?} 
\end{enumerate}

Overall, we make four contributions. First, we establish A/B test simulation via structured questions as a task and provide the first systematic comparison of question formats for eliciting agent preferences. Second, we show that domain alignment of persona source data drives simulation quality, with public behavioral data rivaling platform-specific personas. Third, we empirically characterize the trade-off between per-persona behavioral depth and population diversity. Fourth, we demonstrate that subsampling enables a 2$\times$ cost reduction with no measurable quality loss.
\section{Related Work}
\label{sec:related_work}

\paragraph{LLM-based user simulation.}

LLMs can reproduce human response distributions across demographic groups~\citep{Argyle:2023:PA, Aher:2023:ICML} and approximate population-level behavioral patterns~\citep{Park:2024:arXiv, Bui:2025:ACL}. 
However, they exhibit systematic biases---positional preferences, sycophancy, majority over-representation~\citep{Bisbee:2024:PA, Dominguez-Olmedo:2024:NeurIPS}---that persona conditioning only partially mitigates~\citep{Hu:2024:ACL, Mansour:2025:REALM_paars}, especially when personas lack grounding in observed data.
\wl{Moreover, personalization can increase affective alignment (emotional validation, hedging) even as its effect on epistemic independence varies by interaction role~\citep{Kelley:2026:arXiv}, and demographic cue–based conditioning yields inconsistent behavior depending on how identity is signaled~\citep{Tonneau:2026:arXiv}.}

\paragraph{A/B test simulation.}
SimAB~\cite{Rieder:2026:arXiv_SimAB} frames A/B testing as a preference task with persona-conditioned agents, but uses entirely synthetic personas generated from variant screenshots, achieving 67\% agreement on 47 tests.  SimGym~\cite{Castelo:2026:arXiv_SimGym} grounds simulation in real traffic traces but requires full session logs and does not decouple persona construction from the simulation task. We differ by (i)~constructing personas from real behavioral data independently of test stimuli, (ii)~studying how persona pool composition affects simulation fidelity, and (iii)~introducing subsampling for cost reduction.

\paragraph{Persona generation from data.}
\citet{Mansour:2025:REALM_paars} induce shopping personas from anonymized behavioral data---clickstream, purchases, and pricing signals---and show that persona-conditioned agents better approximate population-level behavior than unprompted models. \citet{Ge:2025:arxiv} scale persona creation to one billion synthetic profiles but without grounding in real user traces. The role-playing literature~\cite{Chen:2024:TMLR, Tseng:2024:EMNLP} finds that richer persona descriptions improve behavioral consistency, yet the source domain and granularity of persona data matter substantially: personas induced from out-of-domain sources rarely outperform basic demographic conditioning unless well-matched to the target population. This motivates our study of how behavioral depth, domain alignment, and population representativeness interact in the A/B testing context.

\paragraph{Evaluation of simulation alignment.}
Evaluating the quality of A/B test simulations can focus on predicting which version performs better (sign-based metrics) or measuring the magnitude of differences (distributional alignment metrics). We adopt a multi-faceted evaluation framework centered on sign-based metrics, including directional accuracy, sign overlap, and sign Bhattacharyya distance, which measure alignment at multiple granularities from individual experiments to distributional patterns. While no standardized evaluation framework currently exists for A/B test simulation specifically, these sign-based approaches address the core question of whether simulations can reliably inform experimental decisions. This evaluation strategy prioritizes capturing the directional patterns that drive real-world A/B testing decisions over achieving perfect numerical precision.
\section{A/B Testing Simulation}
\label{sec:a_b_sim}

In this section, we describe our simulation framework for predicting A/B test outcomes using persona-conditioned LLM agents. We first present the overall pipeline architecture, then detail the three core components: question design for eliciting agent preferences, data-driven persona construction, and population sampling strategies for cost-efficient simulation.

\subsection{Problem Formulation}
\label{sec:problem}

\paragraph{A/B testing.}
A/B testing compares a control group ($C$) and treatment group ($T$) on metrics such as click-through rate (CTR). The experimental framework tests the null hypothesis $H_0: \mu_t = \mu_c$ against the alternative $H_1: \mu_t \neq \mu_c$, where $\mu_t$ and $\mu_c$ represent the true treatment and control CTRs. The design requires pre-specifying Type I error $\alpha$ (typically 0.05) and Type II error $\beta$, which together with the minimum detectable effect determine required sample sizes.

Upon completion, the experiment yields an observed effect size $\hat{\delta} = \hat{\mu}_t - \hat{\mu}_c$. By the Central Limit Theorem, this estimator follows:
$$\hat{\delta} \sim \mathcal{N}\left(\delta_{true}, \frac{\sigma_t^2}{n_t} + \frac{\sigma_c^2}{n_c}\right)$$

For simulation purposes, we establish ground truth as the sampling distribution of the estimator, centered on the observed effect $\mathcal{N}\left(\hat{\delta}_{obs}, \frac{\hat{\sigma}_t^2}{n_t} + \frac{\hat{\sigma}_c^2}{n_c}\right)$. This distribution represents what would be observed across repeated identical experiments and serves as the benchmark for evaluating agent predictions. It captures both the best available estimate of the true effect ($\hat{\delta}_{obs}$) and the appropriate level of uncertainty due to sampling variability.

\paragraph{Simulated A/B testing.} 
We consider A/B testing tasks where control $C$ and treatment $T$ variants are visual stimuli such as advertisement designs or website layouts, evaluated on metrics like click-through rate. The objective is to predict the distribution of the treatment effect size that would emerge from human population testing. To simulate this without requiring actual human participants, we employ LLM-based agents operating under distinct personas. Each persona provides contextual background for an agent to role-play when presented with $C$ versus $T$, generating responses that approximate different human user segments. The simulation outputs a predicted normal distribution of the mean effect size, characterized by estimated mean and standard error (SE). Simulation quality is measured by how closely this predicted distribution matches the true effect size distribution obtained from actual human A/B testing.

\subsection{Framework Overview}
\label{sec:framework}
Our framework consists of three stages:

\paragraph{Persona construction.}
Each persona is constructed from anonymized, aggregated behavioral signals. 
An LLM processes these signals into a structured persona 
(inferred demographics, behavioral summaries, engagement patterns, synthesized narratives, etc.) that serves as conditioning context during simulation. Persona pool construction trades off behavioral depth against population diversity; subsampling strategies reduce redundancy for cost-efficient inference (Sections~\ref{sec:personas},~\ref{sec:sampling}).\footnote{Each persona is a synthetic profile generated from anonymized behavioral signals. While grounded in real activity patterns, it is not intended to represent or re-identify any individual. Inferred demographic attributes serve as statistical conditioning variables for population-level simulation, not as predictions about specific users.}
  
\paragraph{Simulation.}
Given an A/B test with control $C$ and treatments ${T_1, \ldots, T_k}$, each represented as a visual stimulus, and a target metric, we present the variants to each persona-conditioned agent via a structured question. The agent is asked to evaluate them with respect to a target metric and return a numeric score for each variant along with free-text reasoning grounded in its persona profile. The question format---how variants are presented and how responses are elicited---is a design choice that affects simulation quality (Section~\ref{sec:question_design}).
  
\paragraph{Aggregation \& evaluation.}
Per-persona scores are aggregated into a predicted distribution of the treatment effect $\mathcal{N}(\hat{\delta}_s,\hat{\sigma}_s)$, where $\hat{\delta_s} = \bar{s}(T) - \bar{s}(C)$, $\bar{s}(\cdot)$ denotes the mean score across all personas, and $\hat{\sigma}_s$ the standard error of the distribution of score differences over the personas. We evaluate sign alignment using three metrics (Section~\ref{sec:metrics}).

\subsection{Question Design}
\label{sec:question_design}

The question design determines how variants are shown and how preferences are elicited. We study four formats varying along two axes---\emph{isolation vs.\ comparison} (whether the agent sees one variant or both) and \emph{binary vs.\ rating} (whether the response is yes/no or a 1--10 score). In \textbf{independent} formats, each variant is shown separately; in \textbf{pairwise} formats, both variants are presented together with order randomized per persona to control positional bias. Rating formats provide finer-grained signal than binary responses. Full question design descriptions are in Appendix~\ref{app:question_design_descr}.

\subsection{Data-Driven Persona Construction}
\label{sec:personas}

We construct personas by processing anonymized behavioral records through an LLM, which synthesizes raw signals into structured profiles capturing inferred demographics, behavioral statistics, activity patterns, and shopping characteristics (see Appendix~\ref{app:implementation_personas} for full list of attributes). A key design choice is the composition of the persona pool. Two competing objectives arise: \emph{behavioral depth}---maximizing the amount of data available per persona to produce richer, more reliable profiles (deep pool)---and \emph{population diversity}---ensuring the pool covers the demographic and behavioral breadth of the target user base (representative pool). These objectives are in tension as highly active users yield richer personas but may not be representative of the broader population.

\subsection{Population Sampling}
\label{sec:sampling}

Running all personas through the simulation is expensive: inference cost scales linearly with pool size, and large pools often contain redundant profiles that contribute minimal additional signal. We study subsampling strategies that select a smaller subset of personas while preserving the distributional properties of the full pool. 
We compare three algorithms: \emph{uniform random sampling}, \emph{Kernel Herding}~\citep{chen2012super}, which greedily selects personas minimizing maximum mean discrepancy to the full pool, and \emph{Greedy Farthest sampling}~\citep{eldar1997farthest}, which iteratively selects the persona farthest from all the previously selected ones.

\subsection{Evaluation Metrics}
\label{sec:metrics}
Let $p = \Phi(\hat{\delta}/\hat{\sigma})$ and $q = \Phi(\hat{\delta}_s/\hat{\sigma}_s)$ denote the probability of positive effect under ground truth and simulation respectively. We report
\begin{itemize}
    \item \textbf{Accuracy (Acc)}: $\mathbb{I}[(p{-}0.5)(q{-}0.5) > 0]$;
    \item \textbf{Sign overlap (SignOv)}: $1 - |p-q|$;
    \item \textbf{Sign Bhattacharyya (SignBC)}: $(\sqrt{pq} + \sqrt{(1{-}p)(1{-}q)})^2$.
\end{itemize}

Acc captures binary correctness; SignOv and SignBC additionally measure confidence calibration. See Appendix~\ref{app:metrics} for detailed interpretation of the metrics.

\section{Experimental Setup}
\label{sec:exp_setup}

We evaluate our simulation framework on a sample of marketing A/B tests from e-commerce platform. 

\subsection{Benchmark Construction}
\label{sec:benchmark}

Our benchmark comprises 40 experimental test pairs evaluating visual design variants, curated from a sample of historical e-commerce interaction data. 
The benchmark reflects a curated experimental sample and should not be interpreted as representative of any specific platform's full user base or operational A/B testing infrastructure.
Each test compares a control widget against one or more treatment variants, 
with estimated treatment directions provided as confidence intervals on the metric difference (Treatment $-$ Control).
These confidence intervals define a Gaussian distribution $\mathcal{N}(\mu, \sigma)$ over the true effect size, from which we derive directional labels. 
We classify each test in terms of the certainty of its ground-truth outcome as: \emph{positive} ($Pr(\delta > 0) > \tau$), \emph{negative} ($Pr(\delta < 0) > \tau$), or \emph{negligible} ($Pr(|\delta| < \epsilon) > \tau$), excluding indecisive cases.
Thresholds $\tau$ and $\epsilon$ are calibrated per test metric to reflect observed effect size distributions \wl{($\tau=0.8$, $\epsilon=0.01$ for CTR; $\tau=0.7$, $\epsilon=0.1$ for subscriptions). The original candidate set contained over 50 CTR tests and 40 subscription tests; applying these thresholds excluded tests with ambiguous ground truth, yielding the final benchmark of 40 tests (20 per metric).}
The benchmark spans two metric types: click-through rate (engagement) and subscriptions (sign-up intent), evaluated with the same pipeline but different question framing.

\subsection{Personas Pool Construction}
\label{sec:persona_pools}

Personas are constructed from anonymized, aggregated behavioral signals from an e-commerce platform via an LLM that generates structured profiles.~\footnote{See Appendix~\ref{app:implementation_personas} for more details on user profile.}
We compare two persona pools (both containing 935 personas). 
\textbf{Deep pool} contains personas sourced from de-identified activity patterns with high engagement levels, yielding dense behavioral histories.
The richness of the source data produces detailed, high-confidence profiles but skews the pool toward power users who may not be representative of the broader population. \textbf{Representative pool} is a set of users sampled via stratified sampling across population segments to ensure diversity. Individual personas carry sparser behavioral data, but the pool collectively covers a wider range of demographic and behavioral profiles. This pool tests whether breadth of coverage can compensate for reduced per-persona depth at the aggregate simulation level.
Additionally, we construct persona pools (each containing 1000 personas) from three public sources---social science survey data~\citep{Kolluri:2025:EMNLP}, movie ratings~\citep{Leone:2024:RottenTomatoes}, and open e-commerce transactions~\citep{Berke:2023:HarvardDataverse}---to serve as baselines for the data-driven approach and to test whether personas based on public data can rival platform-specific personas.

\subsection{Implementation Details}
\label{sec:implementation}

All experiments use Claude Sonnet 4.5~\citep{Anthropic:2025:Sonnet4.5} via cloud-based batch inference. Persona generation uses two sequential LLM flows (demographics + behavioral narratives) with structured output schemas. Simulation uses temperature 0 for reproducibility, JSON-constrained responses via assistant prefilling, and per-persona randomization of test variant order. \wl{Each agent produces a structured reasoning field grounded in its persona profile before outputting numeric scores, functioning as chain-of-thought prompting (see Appendix~\ref{app:implementation} for full prompt templates and other technical details).}
\section{Results}

We present results organized around the research questions from Section~\ref{sec:intro}, evaluating question design (RQ1), persona data sources (RQ2), the depth--diversity trade-off in persona pool construction (RQ3), and population sampling efficiency (RQ4). In all the following, to assess statistical significance between simulation variants, we apply one-sided paired $t$-tests with $\alpha{=}0.05$.

\subsection{Question Design Comparison}

\begin{table}[t]
    \centering
    \small
    \setlength{\tabcolsep}{3pt}
    \begin{tabular}{l cc cc}
        \toprule
        \multirow{2}{*}{\textbf{Question Design}} & \multicolumn{2}{c}{\textbf{CTR}} & \multicolumn{2}{c}{\textbf{Subscriptions}} \\
        \cmidrule(lr){2-3} \cmidrule(lr){4-5}
        & Acc & SignOv & Acc & SignOv \\
        \hline
        Indep. binary  & 0.40{$\pm$.11} & 0.51{$\pm$.09} & 0.40{$\pm$.11} & 0.56{$\pm$.07} \\
        Pairwise binary     & 0.45{$\pm$.11} & 0.50{$\pm$.09} & \textbf{0.80{$\pm$.08}} & \textbf{0.73{$\pm$.07}} \\
        Indep. rating  & 0.40{$\pm$.11} & 0.52{$\pm$.10} & 0.45{$\pm$.11} & 0.47{$\pm$.08} \\
        Pairwise rating     & \textbf{0.75{$\pm$.10}} & \textbf{0.68{$\pm$.09}} & \textbf{0.80{$\pm$.09}} & 0.72{$\pm$.07} \\
        \bottomrule
    \end{tabular}
    \caption{Question design impact on simulation accuracy comparing independent (Indep.),  pairwise, binary and rating combinations (mean $\pm$ standard error (SE), $n{=}20$ tests per use case). 
    }\label{tab:results:question_designs}
\end{table}

This section addresses \emph{\textbf{RQ1}: the impact of question format on simulation accuracy} (Table~\ref{tab:results:question_designs}).
Pairwise rating consistently outperforms alternatives across all metrics (one-sided paired $t$-tests, $\alpha{=}0.05$). Independent formats perform poorly, suggesting agents struggle to calibrate scores without comparative context. 
Pairwise comparison provides a natural reference point that reduces anchoring, while numeric scales capture preference intensity beyond binary choices. The binary pairwise format achieves strong results on subscription tests but fails on CTR, indicating that optimal design depends on metric type.
\wl{On the combined benchmark ($n=40$), pairwise rating significantly outperforms independent binary on Acc and SignOv, significantly outperforms independent rating on all three metrics, and significantly outperforms pairwise binary on Acc and SignBC. The gap with pairwise binary on SignOv does not reach significance.}

\paragraph{Takeaway:} Pairwise rating is the most effective question format, achieving 0.75 accuracy on CTR and 0.80 on subscription tests. All subsequent experiments use this format.

\subsection{Synthetic vs Data-Driven Personas}
\label{sec:external-results}

\begin{table}[t]
  \centering
  \small
  \setlength{\tabcolsep}{3pt}
  \begin{tabular}{l cc cc}
      \toprule
      \multirow{2}{*}{\textbf{Persona Source}} & \multicolumn{2}{c}{\textbf{CTR}} & \multicolumn{2}{c}{\textbf{Subscriptions}} \\
      \cmidrule(lr){2-3} \cmidrule(lr){4-5}
       & Acc & SignOv & Acc & SignOv \\
      \midrule
      Survey data          & 0.60{$\pm$.11} & 0.58{$\pm$.10} & 0.75{$\pm$.10} & 0.68{$\pm$.07} \\
      Rotten Tomatoes      & 0.65{$\pm$.11} & 0.63{$\pm$.09} & 0.60{$\pm$.11} & 0.57{$\pm$.08} \\
      Open e-comm.      & 0.70{$\pm$.11} & 0.67{$\pm$.09} & \textbf{0.90{$\pm$.07}} & \textbf{0.77{$\pm$.06}} \\
      \hline
      Proprietary & \textbf{0.75{$\pm$.10}} & \textbf{0.68{$\pm$.09}} & 0.80{$\pm$.09} & 0.72{$\pm$.07} \\
      \bottomrule
  \end{tabular}
  \caption{Simulation accuracy across persona sources (mean $\pm$ SE, $n{=}20$ tests per use case). Full results in Appendix~\ref{app:full_tables}.}\label{tab:results:synthetic_personas}
\end{table}

This section addresses \emph{\textbf{RQ2}: whether data-driven personas outperform synthetic personas} (Table~\ref{tab:results:synthetic_personas}).
Personas constructed from platform behavioral data achieve competitive results on CTR tests (0.75 accuracy, 0.68 SignOv) and competitive performance on subscription tests. Among the external persona sources, open e-commerce data performs best and surpasses platform data on subscription tests (0.90 vs.\ 0.80 accuracy), likely due to domain alignment---e-commerce browsing and purchasing signals are directly relevant to evaluating widget engagement and subscription intent. Rotten Tomatoes personas, grounded in entertainment preferences, show reasonable performance on CTR tests (0.65) but degrade on subscription tests (0.60), suggesting that out-of-domain behavioral data provides insufficient signal for metric-specific predictions. 
Survey-based personas perform moderately without excelling on either metric.

\paragraph{Takeaway:} Domain alignment is crucial for personas effectiveness. In-domain behavioral data achieves 0.70--0.90 accuracy, while out-of-domain sources drop to 0.60--0.75. Public e-commerce data can rival platform-specific personas.

\subsection{Persona Pool Comparison}

\begin{table}[t]
    \centering
    \small
    \setlength{\tabcolsep}{3pt}
    \begin{tabular}{l cc cc}
        \toprule
        \multirow{2}{*}{\textbf{Persona Pool}} & \multicolumn{2}{c}{\textbf{CTR}} & \multicolumn{2}{c}{\textbf{Subscriptions}} \\
        \cmidrule(lr){2-3} \cmidrule(lr){4-5} & Acc & SignOv & Acc & SignOv \\
        \midrule
        Deep           & \textbf{0.75{$\pm$.10}} & 0.68{$\pm$.09} & \textbf{0.80{$\pm$.09}} & 0.72{$\pm$.07} \\
        Representative & 0.60{$\pm$.11} & \textbf{0.69{$\pm$.09}} & \textbf{0.80{$\pm$.09}} & \textbf{0.73{$\pm$.07}} \\
        \bottomrule
    \end{tabular}
    \caption{Simulation accuracy across persona pools with large purchases number (Deep) vs representative (mean $\pm$ SE, $n{=}20$ tests per use case). Full results in Appendix~\ref{app:full_tables}.}\label{tab:results:personas}
\end{table}

\begin{figure}[t]
      \centering
      \hspace*{-3mm}%
      \includegraphics[width=1.07\linewidth]{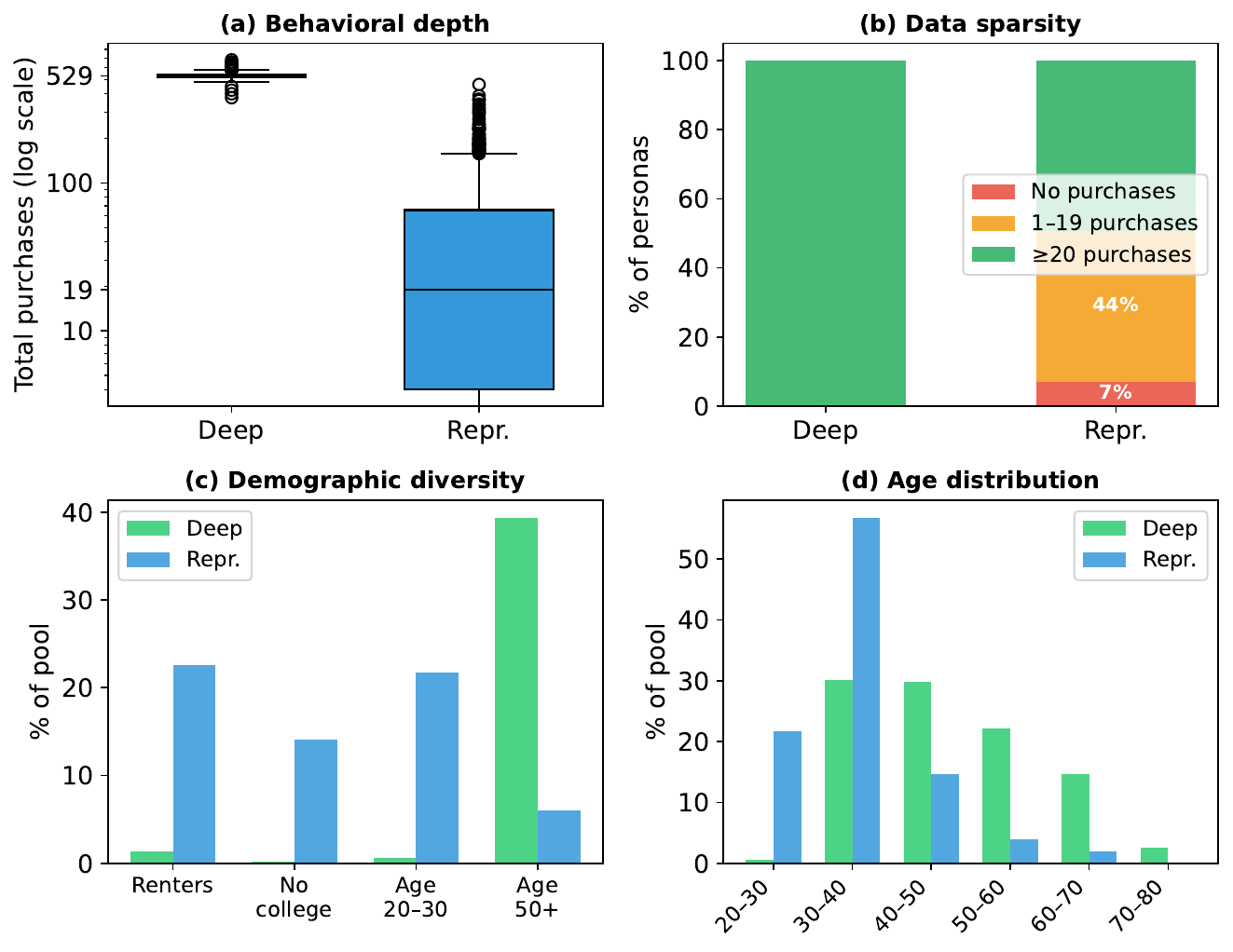}
      \caption{Persona pool comparison: (a) behavioral depth (total \# transactions), (b) data sparsity, (c) demographic diversity on key under-represented segments, and (d) age distribution with concentration bias.}
      \label{fig:pool-analysis}
  \end{figure}

This section addresses \emph{\textbf{RQ3}: the trade-off between per-persona behavioral depth and population-level diversity} (Table~\ref{tab:results:personas}).
The deep pool significantly outperforms the representative pool on CTR accuracy (0.75 vs.\ 0.60, one-sided paired $t$-test, $\alpha{=}0.05$), while on subscription tests there is no statistically significant difference on any metric (0.80 accuracy for both).
Despite the accuracy gap on CTR tests, the distributional metrics differ only marginally (SignOv 0.68 vs.\ 0.69), suggesting that demographic coverage largely compensates for reduced behavioral depth at the population level.
The deep pool is characterized by dense interaction histories vs. sparse signals for the representative pool (Figure~\ref{fig:pool-analysis}; panel a), with a substantial portion of representative personas having minimal interaction data (panel b).
\wl{We hypothesize that} for sparse personas, the LLM lacks sufficient behavioral grounding and defaults to generic reasoning rather than user-specific preferences, explaining the \wl{CTR accuracy gap}.
The representative pool compensates through demographic diversity (panel c): 
it introduces substantially more renters, non-college-educated users, and younger demographics that are nearly absent from the deep pool, which skews toward homeowning, educated users.
Both pools exhibit complementary demographic gaps (panel d): the representative pool over-concentrates in the 30–40 age range with minimal older representation, while the deep pool is more spread but still under-represents younger users.
These gaps likely arise from LLM inference biases during persona generation rather than sampling limitations, and represent a primary lever for future improvement.~\footnote{These distributions reflect our sampled experimental cohort and should not be interpreted as representative of any specific platform's full user base.}

\paragraph{Takeaway:} Behavioral depth yields a statistically significant advantage on CTR accuracy (0.75 vs.\ 0.60, $p{<}0.05$), but demographic diversity fully compensates on subscription tests (no significant difference on any metric). The remaining gap points to persona quality---sparse-data handling and demographic calibration.

\subsection{Population Sampling Efficiency}

\begin{table}[t]
    \centering
    \small
    \setlength{\tabcolsep}{3pt}
    \begin{tabular}{l cc cc}
        \toprule
        & \multicolumn{2}{c}{\textbf{CTR}} & \multicolumn{2}{c}{\textbf{Subscriptions}} \\
        \cmidrule(lr){2-3} \cmidrule(lr){4-5}
        \textbf{Algorithm} & Acc & SignOv & Acc & SignOv \\
        \midrule
        Random          & 0.74 & 0.68 & 0.81 & 0.70 \\
        Kernel Herding  & 0.73 & 0.67 & \textbf{0.81} & \textbf{0.70} \\
        Greedy Farthest & \textbf{0.75} & \textbf{0.68} & 0.80 & 0.69 \\
        \midrule
        All (935)       & 0.75 & 0.68 & 0.80 & 0.72 \\
        \bottomrule
    \end{tabular}
    \caption{Subsampling performance at 500 personas (mean $\pm$ SE across 1000 trials, all SEs are below 0.005 hence omitted for brevity). Full results in Appendix~\ref{app:full_tables}.}\label{tab:results:subsampling}
\end{table}

This section addresses \emph{\textbf{RQ4}: the trade-off between persona pool size reduction via subsampling and preserving simulation quality}. We compare three subsampling strategies---Random, Greedy Farthest, and Kernel Herding---for selecting representative personas from a pool of 935 personas.
These algorithms operate on vector representations and we apply them to persona embeddings computed using Qwen2.5-32B-Instruct \citep{qwen2.5}. Detailed algorithm descriptions are provided in Appendix~\ref{app:subsampling}.

\noindent
Table~\ref{tab:results:subsampling} presents the performance metrics when sampling 500 personas from the pool of 935 using each algorithm.
Results are averaged over 1000 independent trials. At 500 personas, Kernel Herding performs best for subscription tests while Greedy Farthest is best for CTR. However, even random sampling achieves strong performance, suggesting reduced persona sets provide robust signals.

\paragraph{Takeaway:} All subsampling strategies preserve near-full-pool accuracy at 500 personas (within 1pp on CTR, matching or exceeding on subscriptions), potentially enabling up to 2$\times$ cost reduction. Pairwise comparison format provides natural robustness, though geometric methods offer consistent modest gains, particularly for sign-based metrics.

\subsection{\wl{Ablation: Component Contributions}}

To isolate the contribution of each framework component, we evaluate four baselines using the pairwise rating format (Table~\ref{tab:results:ablation_baselines}). Two baselines use the full deep persona pool (935 personas), modifying the persona representation or prompting strategy: \textit{demographics only} strips the persona to demographic attributes (age, gender, education, income, etc.); \textit{no reasoning} uses the full persona but instructs the model to output only scores without reasoning (no chain-of-thought requirement). Two baselines test minimal-conditioning scenarios: \textit{generic shopper} conditions the model on a single fixed persona applied uniformly across all tests;\footnote{Generic shopper prompt: \emph{``A 35-year-old suburban homeowner with moderate income who shops online regularly for household essentials, electronics, and clothing.''}} \textit{no persona} removes persona conditioning entirely.

\begin{table}[t]
  \centering
  \small
  \setlength{\tabcolsep}{2.5pt}
  \hspace*{-1mm}
  \begin{tabular}{l cc cc}
      \toprule
      \multirow{2}{*}{\textbf{Configuration}} & \multicolumn{2}{c}{\textbf{CTR}} & \multicolumn{2}{c}{\textbf{Subscriptions}} \\
      \cmidrule(lr){2-3} \cmidrule(lr){4-5}
      & Acc & SignOv & Acc & SignOv \\
      \hline
      Pairwise rating   & \textbf{0.75{$\pm$.10}} & \textbf{0.68{$\pm$.09}} & \textbf{0.80{$\pm$.09}} & \textbf{0.72{$\pm$.07}} \\
      Demographics & 0.65{$\pm$.11} & 0.57{$\pm$.08} & 0.30{$\pm$.11} & 0.39{$\pm$.07} \\
      No reasoning      & 0.60{$\pm$.11} & 0.60{$\pm$.09} & \textbf{0.80{$\pm$.09}} & \textbf{0.73{$\pm$.06}} \\
      Generic shopper  & 0.40{$\pm$.11} & 0.44{$\pm$.10} & 0.60{$\pm$.11} & 0.64{$\pm$.08} \\
      No persona       & 0.45{$\pm$.11} & 0.49{$\pm$.10} & 0.65{$\pm$.11} & 0.63{$\pm$.08} \\
      \bottomrule
  \end{tabular}
    \caption{
  Ablation studies comparing the best system (pairwise rating) vs demographics only and no reasoning using 935 personas (mean $\pm$ SE, $n{=}20$ tests per use case). We include "non-simulation" baselines that do not use personas (single LLM call with generic shopper and no persona). Full results in Appendix~\ref{app:full_tables}.
  }\label{tab:results:ablation_baselines}
\end{table}

Full behavioral personas achieve +10--35pp accuracy lift on CTR and +15--50pp on subscriptions over the baselines. Demographics-only personas degrade substantially on subscriptions (0.30 vs.\ 0.80), indicating that behavioral profiles are critical for higher-salience decisions. Removing reasoning drops CTR accuracy from 0.75 to 0.60 while subscription accuracy is preserved (0.80), suggesting that CTR prediction benefits from explicit persona-grounded deliberation. A single generic persona does not outperform the unconditioned LLM (0.40 vs.\ 0.45 on CTR, 0.60 vs.\ 0.65 on subscriptions), confirming that population diversity---not mere persona framing---drives the improvement.

\paragraph{Takeaway:} Persona conditioning with behavioral data provides substantial lift over simpler alternatives. Behavioral depth beyond demographics is most critical for subscription tests, while chain-of-thought reasoning primarily benefits CTR prediction. Population diversity is essential---a single generic persona offers no advantage over no persona at all.

\subsection{\wl{Multi-Model Validation}}

To assess whether our findings generalize beyond a single LLM, we replicate the question design comparison (Table~\ref{tab:results:question_designs}) with two additional models: Claude Haiku \citep{Anthropic:2025:Haiku.5} and Claude Opus \citep{Anthropic:2025:Opus4.5} (see Table~\ref{tab:results:multi_model}). All other settings remain identical (deep persona pool, temperature 0, per-persona randomization of test variant order).

\begin{table}[t]
  \centering
  \small
  \setlength{\tabcolsep}{2.5pt}
  \begin{tabular}{l l ccc}
      \toprule
      \textbf{Model} & \textbf{Quest. Design} & Acc & SignBC & SignOv \\
      
      \hline
      \multirow{4}{*}{Haiku}
      & Indep. binary  & 0.55$\pm$0.08 & 0.62$\pm$0.06 & 0.59$\pm$0.06 \\
      & Pairwise binary     & \textbf{0.75$\pm$0.07} & 0.65$\pm$0.06 & \textbf{0.64$\pm$0.06} \\
      & Indep. rating  & 0.65$\pm$0.08 & \textbf{0.66$\pm$0.06} & \textbf{0.64$\pm$0.06} \\
      & Pairwise rating     & 0.70$\pm$0.07 & 0.65$\pm$0.06 & \textbf{0.64$\pm$0.06} \\
      \hline
      \multirow{4}{*}{Sonnet}
      & Indep. binary  & 0.40$\pm$0.08 & 0.59$\pm$0.06 & 0.54$\pm$0.06 \\
      & Pairwise binary     & 0.62$\pm$0.07 & 0.62$\pm$0.06 & 0.61$\pm$0.06 \\
      & Indep. rating  & 0.43$\pm$0.08 & 0.52$\pm$0.06 & 0.49$\pm$0.06 \\
      & Pairwise rating     & \textbf{0.78$\pm$0.07} & \textbf{0.71$\pm$0.06} & \textbf{0.70$\pm$0.06} \\
      \hline
      \multirow{4}{*}{Opus}
      & Indep. binary  & 0.55$\pm$0.08 & 0.56$\pm$0.06 & 0.52$\pm$0.06 \\
      & Pairwise binary     & \textbf{0.75$\pm$0.07} & \textbf{0.73$\pm$0.05} & \textbf{0.72$\pm$0.05} \\
      & Indep. rating  & 0.50$\pm$0.08 & 0.50$\pm$0.07 & 0.47$\pm$0.07 \\
      & Pairwise rating     & 0.60$\pm$0.08 & 0.64$\pm$0.06 & 0.63$\pm$0.06 \\
      \bottomrule
  \end{tabular}
  \caption{
  Robustness across models with different capabilities to various question designs (combined benchmark, $n=40$). Values are mean $\pm$ SE. Statistical significance via one-sided paired $t$-tests ($\alpha=0.05$).
  }\label{tab:results:multi_model}
\end{table}

Across all three models, pairwise settings consistently outperform independent settings. On the combined benchmark, the difference between pairwise rating and pairwise binary is not statistically significant for any model, though pairwise rating achieves the highest combined accuracy for Sonnet 4.5. The pairwise binary format performs particularly well on Haiku and Opus, driven by strong subscription test performance. This consistency confirms that the advantage of pairwise presentation---forcing comparative evaluation rather than uncalibrated absolute scoring---is a robust finding independent of the specific model used. The pairwise rating format additionally provides more interpretable, granular responses, justifying its use as the default configuration.

\paragraph{Takeaway:} Pairwise formats consistently outperform independent formats across all three models. The choice between pairwise rating and pairwise binary is model-dependent, but pairwise rating offers greater interpretability at comparable accuracy.
\section{Enabling Reproducible Research}
\label{sec:reproducible_research}

\begin{figure*}[t]
    \centering
    \includegraphics[width=0.9\linewidth]{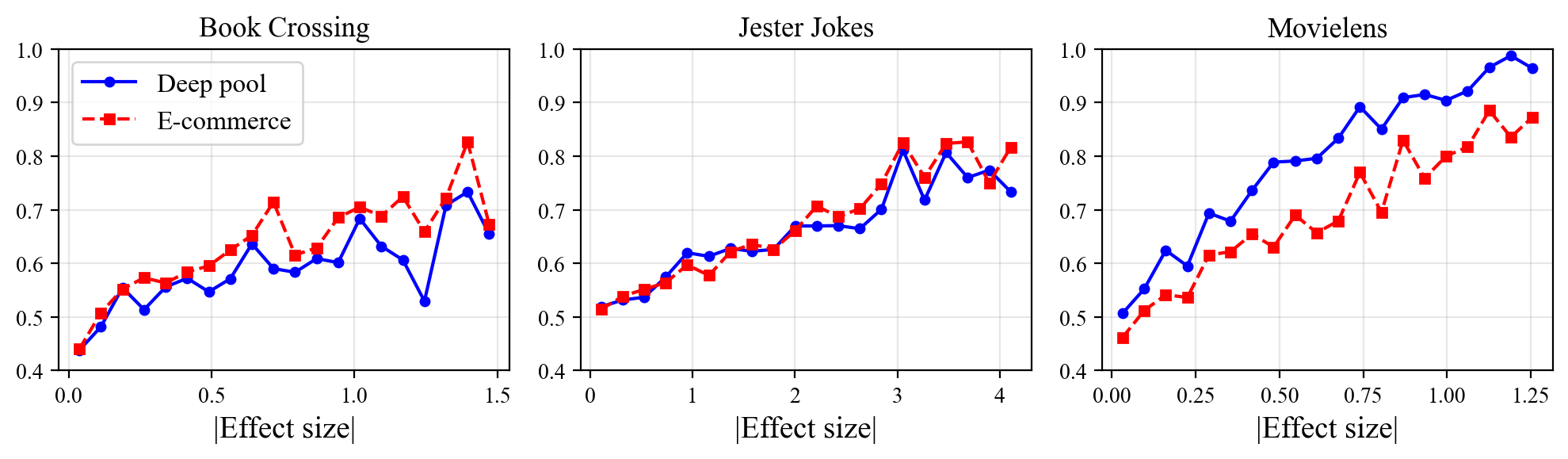}
    \caption{Accuracy of A/B test prediction vs absolute effect size, comparing our deep pool of personas against public open e-commerce personas. The trends look similar, hence supporting reproducibility and future work without access to proprietary data.}
    \label{fig:accuracy_three_datasets}
\end{figure*}

While our benchmark relies on proprietary A/B test outcomes, the methodology is fully reproducible with public resources. Any preference dataset with ground-truth rankings---movie ratings, product reviews, website redesign preferences---can be reframed as a simulated A/B test by treating item pairs as control/treatment variants and user preferences as ground-truth effect directions.

\subsection{Evaluation on Public Datasets}

To demonstrate reproducibility, we evaluate our persona-based A/B test simulation framework on three public benchmarks: Book Crossing~\citep{ziegler2005improving}, Jester Jokes~\citep{goldberg2001eigentaste}, and MovieLens~\citep{harper2015movielens}.

\paragraph{Dataset preparation.}
We sample 100 items from each dataset: books rated on a 1--10 integer scale (Book Crossing), jokes rated on a $-10$ to $10$ scale with 0.5 increments (Jester), and movies rated on a 0.5--5.0 scale with 0.5 increments (MovieLens). For each item, we compute ground-truth statistics from real user ratings: mean rating ($\mu_{GT}$) and standard error of the mean ($\sigma_{GT}$).

\paragraph{Pairwise comparison framework.}
For each pair of items $(A, B)$ within a dataset, we construct an A/B test scenario. The ground-truth effect is the difference in mean ratings, $\Delta_{GT} = \mu_A - \mu_B$, with uncertainty $\sigma_{\Delta,GT} = \sqrt{\sigma_A^2 + \sigma_B^2}$. To obtain the predicted effect, we query each persona to rate both items, compute per-persona rating differences, and aggregate to obtain the empirical mean $\Delta_{pred}$ and standard error $\sigma_{\Delta,pred}$. This yields $\binom{100}{2} = 4{,}950$ pairwise comparisons per dataset.

\paragraph{Persona configurations.}
We evaluate two persona configurations: the deep persona pool from our main experiments and a public e-commerce persona set based on open shopping profiles~\citep{Berke:2023:HarvardDataverse}. This allows us to assess whether the framework's effectiveness depends on proprietary behavioral data.

\paragraph{Evaluation metrics.}
For each pairwise comparison, we compute the probability that the treatment is better than the reference from both ground-truth and simulation data: $p_{GT} = P(\Delta_{GT} > 0)$ and $p_{pred} = P(\Delta_{pred} > 0)$, assuming Gaussian distributions. We then measure directional accuracy as a function of absolute effect size.

\subsection{Results and Implications}

Figure~\ref{fig:accuracy_three_datasets} shows directional accuracy as a function of absolute effect size across the three public datasets, evaluated with both persona configurations. Each point represents the average accuracy within effect size bins.

Several patterns emerge. First, accuracy increases monotonically with effect size across all datasets and both persona configurations, confirming the intuitive expectation that larger differences between items are easier to predict. Second, even at moderate effect sizes, both persona types achieve accuracy well above chance (0.5), demonstrating that the simulation framework produces meaningful signal on fully public data. Third, the two persona configurations exhibit consistent behavior despite being constructed from entirely different data sources, suggesting that the framework's effectiveness is not contingent on access to proprietary behavioral data.

These results establish a concrete path for reproducibility: researchers can construct both the persona pool and the evaluation benchmark from permissive public datasets, enabling direct extension and comparison without access to proprietary experimentation infrastructure. This evaluation framework allows researchers to benchmark A/B test simulation approaches, reproduce experiments from our paper, and develop further research in this direction using publicly available resources.

\section{Discussion}

\paragraph{\mf{Potential applications}.}
With current accuracy levels, the proposed framework cannot fully replace human A/B tests---but it does not need to. 
A potential application could be a pre-screening tool that filters clearly inferior treatment candidates before they consume traffic and prioritizes the experiments by ranking proposed changes by predicted impact.
With batch inference, a full simulation could complete in hours at a fraction of a multi-week experiment cost, potentially enabling teams to explore a broader design space without proportionally increasing experimentation overhead.

\paragraph{When to trust simulation.}
\mf{Our experiments show that simulation results are most reliable when the underlying effect size is large, as is typical of high-salience decisions. Conversely, the framework is least trustworthy for near-zero-effect tests where small perturbations flip the predicted direction. This suggests that outputs could be more useful as a ranking signal rather than a binary decision criterion.}

\paragraph{Cost--quality trade-off.}
Subsampling to 500 personas performs within 1pp of the full pool across all metrics for subscription tests, and matches the full pool on CTR. Even at $n{=}100$, all algorithms remain competitive (0.77--0.80 CTR, 0.82--0.86 subscriptions), making routine simulation viable under constrained budgets (more details in Appendix~\ref{app:subsampling}).

\paragraph{Data requirements.}
Domain alignment matters more than data volume or source exclusivity. Public e-commerce data rivals platform-specific personas (Table~\ref{tab:results:synthetic_personas}), lowering adoption barriers. However, below ${\sim}20$ recorded transactions, simulation degrades as the LLM defaults to generic reasoning.
\section{Conclusions}

We presented a framework for simulating A/B tests using LLM agents conditioned on data-driven personas constructed from real behavioral data. Through systematic evaluation on 40 A/B tests across two metric types, we established four findings: (i)~pairwise rating is the most effective question format for eliciting agent preferences, achieving 0.75--0.80 directional accuracy; (ii)~domain alignment of persona source data is the primary driver of simulation quality, with publicly available e-commerce data rivaling platform-specific personas; (iii)~demographic diversity can largely compensate for reduced per-persona behavioral depth at the aggregate level; and (iv)~Kernel Herding enables efficient subsampling to 500 personas with negligible quality loss. 
These results suggest that data-driven persona simulation could serve as a viable tool for pre-screening A/B test candidates, potentially reducing wasted experimentation traffic while maintaining directional accuracy that may be sufficient for prioritization decisions.

\section*{Limitations}

Our framework evaluates agents on isolated screenshots rather than full page contexts, removing situational factors (browsing intent, session history, surrounding content) that influence real user decisions. The benchmark is limited to 40 tests from a single e-commerce domain and two metric types; generalization to other domains, metric types, or multi-step user journeys remains untested. Persona demographics are inferred by the LLM from behavioral signals rather than self-reported, introducing systematic biases that may distort population-level predictions. LLM positivity bias and anchoring effects likely produce systematically optimistic treatment evaluations, a tendency that our sign-based metrics partially mask when both control and treatment are equally inflated.
\wl{Persona conditioning may also amplify affective alignment (increasing emotional validation and hedging in model responses) even when epistemic independence is preserved~\citep{Kelley:2026:arXiv}. Our pairwise format partially mitigates this, as uniform positivity bias cancels when computing relative treatment effects. Additionally, inferred demographics depend on how identity is cued through behavioral signals; different cue types can yield inconsistent demographic conditioning~\citep{Tonneau:2026:arXiv}, and our pipeline relies on a single cue modality (aggregated purchase behavior). All experiments use a single LLM (Claude Sonnet 4.5); while we validate consistency across Claude Haiku and Opus, generalization to non-Anthropic models remains untested. Few-shot prompting was not explored, as no ground-truth persona $\rightarrow$ score pairs exist to serve as demonstrations without risking score anchoring.}
Finally, the benchmark ground-truth labels are not publicly released, limiting external reproducibility; however, the methodology is fully reproducible with any preference dataset, as we demonstrate with public data sources that produce competitive results.

\section*{Ethical Considerations}
All behavioral data used for persona construction is anonymized and de-identified prior to processing; no personally identifiable information is included in the pipeline. Inferred demographic attributes (age, gender, income) are generated by the LLM as approximate statistical priors — they may not reflect users' actual characteristics and are not validated against self-reported data. These attributes serve solely as conditioning signals for population-level simulation and are not attributed to individuals or used for any downstream profiling or targeting purpose. The persona generation pipeline does not retain explicit links to source records.

\bibliography{emnlp2026}

\appendix
\section{Question Design: Extended Analysis}
\label{app:question_design}

\subsection{Designs Description}
\label{app:question_design_descr}

We frame A/B test simulation as a structured question task: each persona-conditioned agent is presented with variant screenshots and asked to evaluate them with respect to a target metric. The question format determines how variants are shown and how preferences are elicited, introducing a design choice that affects both the cognitive framing for the LLM and the resulting simulation quality. We study four question designs:
\begin{itemize}[leftmargin=1em, itemsep=0pt, parsep=0pt, topsep=3pt]
    \item \textbf{Independent binary}: Each variant is shown separately, and the agent decides whether they would engage or not, giving only a binary signal. Responses to different variants are fully independent and unanchored, closely mimicking real-world user behavior.
    \item \textbf{Independent rating}: Similar to the binary setting, each variant is shown in isolation, and here the agent is asked to rate it on a 1--10 Likert scale without seeing alternatives. This provides greater response granularity and signal strength.
    \item \textbf{Pairwise binary}: Each question presents a pair of variants, asking the agent whether they would engage with each one. This enables direct comparison between the two options.
    Variant order is randomized per persona to mitigate positional bias.
    \item \textbf{Pairwise rating}: As in pairwise binary, every pair of variants is presented simultaneously, but agents rate each option on a 1--10 Likert scale rather than making binary engagement decisions.
\end{itemize}

\noindent
The formats differ in whether the agent sees variants in isolation or in context of alternatives. This choice interacts with the LLM's tendency toward anchoring and positional bias, making question design a non-trivial factor in simulation accuracy.

\section{Evaluation Metrics: Intuition via Toy Examples}
\label{app:metrics}

The human A/B test outcome represents ground truth as a normal distribution $\mathcal{N}(\hat{\delta}_{\text{rel}},\hat{\sigma}_{\text{rel}}^2)$, while the simulation produces a predicted distribution $\mathcal{N}(\hat{\delta}_s,\hat{\sigma}_s^2)$. This predicted distribution is obtained by computing the relative effect predicted by each persona, then calculating the mean and standard error across these persona-level relative scores. Since we focus on sign alignment (i.e., predicting which version drives higher engagement), we evaluate whether both distributions agree on the direction of the effect. Let $p = \Phi\left(\frac{\hat{\delta}_{\text{rel}}}{\hat{\sigma}_{\text{rel}}}\right)$ denote the probability that the ground truth relative effect is positive, and $q = \Phi\left(\frac{\hat{\delta}_s}{\hat{\sigma}_s}\right)$ the probability that the simulated effect is positive, where $\Phi$ is the standard normal CDF. We compare $p$ and $q$ using three complementary metrics:

\begin{itemize}[leftmargin=1em, itemsep=0pt, parsep=0pt, topsep=3pt]
    \item \textbf{Simple accuracy}: $\mathbb{I}[(p-0.5)(q-0.5) > 0]$, which equals 1 if and only if $\hat{\delta}$ and $\hat{\delta}_s$ have the same sign.
    \item \textbf{Sign overlap}: $\min(p,q) + \min(1-p,1-q) = 1-|p-q|$. This smooth metric represents the probability mass where both distributions agree on sign. It equals 1 when $p=q$ (perfect agreement) and 0 when one probability is 0 and the other is 1 (complete disagreement). It also corresponds to 1 minus the total variation distance between the corresponding Bernoulli distributions.
    \item \textbf{Sign Bhattacharyya coefficient}: $\text{BC}^2 = (\sqrt{pq} + \sqrt{(1-p)(1-q)})^2$. This metric also achieves its maximum of 1 when $p=q$ but penalizes disagreement differently than sign overlap. While sign overlap penalizes linearly in $|p-q|$, the Bhattacharyya coefficient provides a more gradual penalty that is less sensitive to small deviations when both $p$ and $q$ are near the boundaries (0 or 1). Note that this metric also equals the squared cosine similarity between $(\sqrt{p}, \sqrt{1-p})$ and $(\sqrt{q}, \sqrt{1-q})$.
\end{itemize}

\begin{figure}[h]
    \centering
    \includegraphics[width=0.95\linewidth]{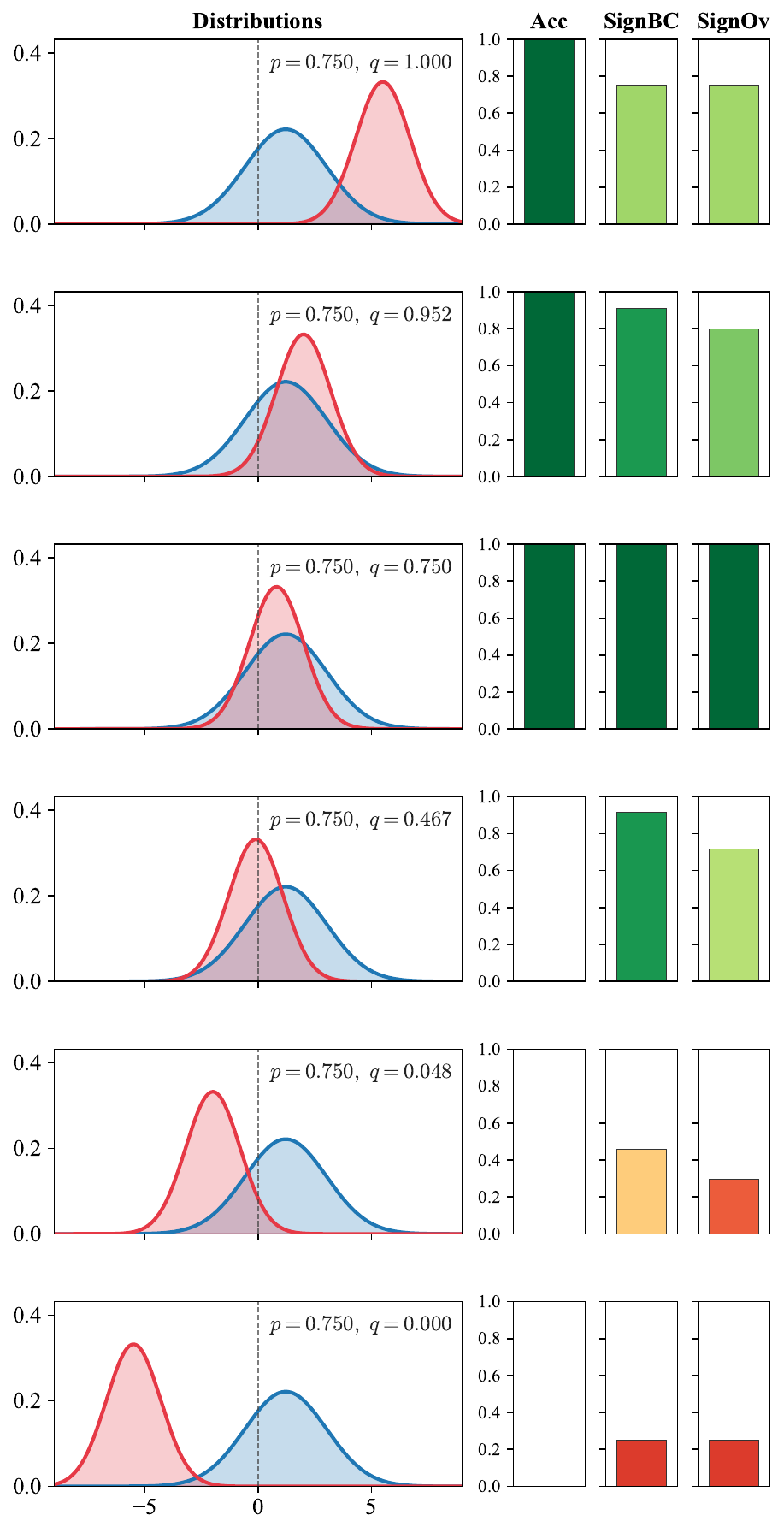}
    \caption{Sign metrics for ground truth and predicted effect size distribution.}
    \label{fig:sign-metrics}
\end{figure}

Figure \ref{fig:sign-metrics} compares the metrics Acc, SignBC, and SignOv for different input Gaussians. As shown in the figure, Acc can be highly noisy, especially when the means of both distributions are near zero. In such cases, small perturbations can flip the sign of either distribution, causing accuracy to change abruptly from 0 to 1 or vice versa. Moreover, accuracy does not account for the uncertainty in either distribution. In contrast, SignOv and SignBC exhibit smoother behavior that varies continuously with respect to the input distributions.

\subsection{Magnitude Alignment Metrics}

In practice, the primary decision derived from an A/B test is whether to \emph{ship} a treatment---a decision that hinges almost entirely on the \emph{direction} (sign) of the effect, not its precise magnitude. A simulation that correctly predicts that variant B increases engagement by \emph{some positive amount} provides actionable guidance, even if it overestimates or underestimates the exact lift. Conversely, a simulation that precisely estimates the magnitude but assigns it the wrong sign leads to an incorrect ship/no-ship decision.

\begin{figure}[t]
    \centering
    \includegraphics[width=0.95\linewidth]{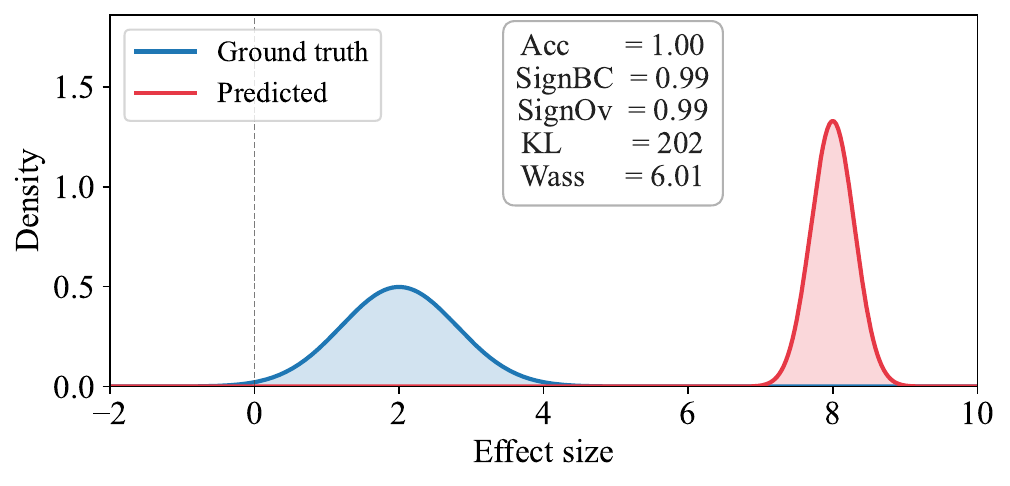}
    \caption{Sign alignment vs magnitude alignment}
    \label{fig:sign_vs_magnitude}
\end{figure}

Classical distributional distance metrics such as the Kullback--Leibler (KL) divergence and the Wasserstein distance measure how closely the predicted effect-size distribution matches the ground truth in its entirety. While appropriate when the full distribution is of interest, they conflate two fundamentally different types of error: \textbf{directional error} (predicting the wrong sign) and \textbf{magnitude error} (predicting the correct sign but with inaccurate magnitude). As illustrated in Figure~\ref{fig:sign_vs_magnitude}, a prediction that correctly identifies a positive effect but overestimates its magnitude receives good scores under the sign metrics but poor ones under KL and Wasserstein distances (the lower the better), despite being operationally correct.

That said, magnitude metrics remain valuable in other settings where precise effect-size estimation is critical---such as cost-benefit analysis---where knowing \emph{how much} matters as much as knowing the direction.

\section{Full Results Tables}
\label{app:full_tables}

Tables~\ref{tab:app:question_designs_full}--\ref{tab:app:ablation_full} report the complete results for the main experiments, including the SignBC metric omitted from the main text due to space constraints, and combined results over the full benchmark ($n{=}40$). All values are mean $\pm$ SE; combined mean = average of CTR and Subscriptions means; combined SE = $\frac{1}{2}\sqrt{\text{SE}_\text{CTR}^2 + \text{SE}_\text{Sub}^2}$.

\begin{table*}[h]
    \centering
    \small
    \setlength{\tabcolsep}{3pt}
    \begin{tabular}{l ccc ccc ccc}
        \toprule
        \multirow{2}{*}{\textbf{Question Design}} & \multicolumn{3}{c}{\textbf{CTR} ($n{=}20$)} & \multicolumn{3}{c}{\textbf{Subscriptions} ($n{=}20$)} & \multicolumn{3}{c}{\textbf{Combined} ($n{=}40$)} \\
        \cmidrule(lr){2-4} \cmidrule(lr){5-7} \cmidrule(lr){8-10}
        & Acc & SignBC & SignOv & Acc & SignBC & SignOv & Acc & SignBC & SignOv \\
        \hline
        Independent binary  & 0.40{\tiny$\pm$.11} & 0.53{\tiny$\pm$.09} & 0.51{\tiny$\pm$.09} & 0.40{\tiny$\pm$.11} & 0.65{\tiny$\pm$.08} & 0.56{\tiny$\pm$.07} & 0.40{\tiny$\pm$.08} & 0.59{\tiny$\pm$.06} & 0.54{\tiny$\pm$.06} \\
        Pairwise binary     & 0.45{\tiny$\pm$.11} & 0.51{\tiny$\pm$.09} & 0.50{\tiny$\pm$.09} & \textbf{0.80{\tiny$\pm$.08}} & 0.73{\tiny$\pm$.07} & \textbf{0.73{\tiny$\pm$.07}} & 0.62{\tiny$\pm$.07} & 0.62{\tiny$\pm$.06} & 0.61{\tiny$\pm$.06} \\
        Independent rating  & 0.40{\tiny$\pm$.11} & 0.54{\tiny$\pm$.10} & 0.52{\tiny$\pm$.10} & 0.45{\tiny$\pm$.11} & 0.50{\tiny$\pm$.08} & 0.47{\tiny$\pm$.08} & 0.43{\tiny$\pm$.08} & 0.52{\tiny$\pm$.06} & 0.49{\tiny$\pm$.06} \\
        Pairwise rating     & \textbf{0.75{\tiny$\pm$.10}} & \textbf{0.69{\tiny$\pm$.09}} & \textbf{0.68{\tiny$\pm$.09}} & \textbf{0.80{\tiny$\pm$.09}} & \textbf{0.74{\tiny$\pm$.07}} & 0.72{\tiny$\pm$.07} & \textbf{0.78{\tiny$\pm$.07}} & \textbf{0.71{\tiny$\pm$.06}} & \textbf{0.70{\tiny$\pm$.06}} \\
        \bottomrule
    \end{tabular}
    \caption{Survey design impact on simulation accuracy (mean $\pm$ SE). Full version of Table~\ref{tab:results:question_designs}.}
    \label{tab:app:question_designs_full}
\end{table*}

\begin{table*}[h]
    \centering
    \small
    \setlength{\tabcolsep}{3pt}
    \begin{tabular}{l ccc ccc ccc}
        \toprule
        \multirow{2}{*}{\textbf{Persona Source}} & \multicolumn{3}{c}{\textbf{CTR} ($n{=}20$)} & \multicolumn{3}{c}{\textbf{Subscriptions} ($n{=}20$)} & \multicolumn{3}{c}{\textbf{Combined} ($n{=}40$)} \\
        \cmidrule(lr){2-4} \cmidrule(lr){5-7} \cmidrule(lr){8-10}
        & Acc & SignBC & SignOv & Acc & SignBC & SignOv & Acc & SignBC & SignOv \\
        \midrule
        Survey data          & 0.60{\tiny$\pm$.11} & 0.59{\tiny$\pm$.10} & 0.58{\tiny$\pm$.10} & 0.75{\tiny$\pm$.10} & 0.71{\tiny$\pm$.07} & 0.68{\tiny$\pm$.07} & 0.68{\tiny$\pm$.07} & 0.65{\tiny$\pm$.06} & 0.63{\tiny$\pm$.06} \\
        Rotten Tomatoes      & 0.65{\tiny$\pm$.11} & 0.64{\tiny$\pm$.09} & 0.63{\tiny$\pm$.09} & 0.60{\tiny$\pm$.11} & 0.59{\tiny$\pm$.08} & 0.57{\tiny$\pm$.08} & 0.62{\tiny$\pm$.08} & 0.61{\tiny$\pm$.06} & 0.60{\tiny$\pm$.06} \\
        Open e-commerce      & 0.70{\tiny$\pm$.11} & 0.68{\tiny$\pm$.10} & 0.67{\tiny$\pm$.09} & \textbf{0.90{\tiny$\pm$.07}} & \textbf{0.77{\tiny$\pm$.06}} & \textbf{0.77{\tiny$\pm$.06}} & \textbf{0.80{\tiny$\pm$.07}} & \textbf{0.73{\tiny$\pm$.06}} & \textbf{0.72{\tiny$\pm$.05}} \\
        \hline
        Platform data (ours) & \textbf{0.75{\tiny$\pm$.10}} & \textbf{0.69{\tiny$\pm$.09}} & \textbf{0.68{\tiny$\pm$.09}} & {0.80{\tiny$\pm$.09}} & {0.74{\tiny$\pm$.07}} & 0.72{\tiny$\pm$.07} & {0.78{\tiny$\pm$.07}} & {0.71{\tiny$\pm$.06}} & {0.70{\tiny$\pm$.06}} \\
        \bottomrule
    \end{tabular}
    \caption{Simulation accuracy across persona sources (mean $\pm$ SE). Full version of Table~\ref{tab:results:synthetic_personas}.}
    \label{tab:app:synthetic_personas_full}
\end{table*}

\begin{table*}[h]
    \centering
    \small
    \setlength{\tabcolsep}{3pt}
    \begin{tabular}{l ccc ccc ccc}
        \toprule
        \multirow{2}{*}{\textbf{Persona Pool}} & \multicolumn{3}{c}{\textbf{CTR} ($n{=}20$)} & \multicolumn{3}{c}{\textbf{Subscriptions} ($n{=}20$)} & \multicolumn{3}{c}{\textbf{Combined} ($n{=}40$)} \\
        \cmidrule(lr){2-4} \cmidrule(lr){5-7} \cmidrule(lr){8-10}
        & Acc & SignOv & SignBC & Acc & SignOv & SignBC & Acc & SignOv & SignBC \\
        \midrule
        Deep           & \textbf{0.70{\tiny$\pm$.10}} & 0.62{\tiny$\pm$.09} & \textbf{0.63{\tiny$\pm$.09}} & \textbf{0.70{\tiny$\pm$.09}} & 0.64{\tiny$\pm$.07} & \textbf{0.67{\tiny$\pm$.07}} & \textbf{0.71{\tiny$\pm$.07}} & 0.63{\tiny$\pm$.06} & \textbf{0.64{\tiny$\pm$.06}} \\
        Representative & 0.55{\tiny$\pm$.11} & \textbf{0.63{\tiny$\pm$.09}} & 0.61{\tiny$\pm$.10} & \textbf{0.70{\tiny$\pm$.09}} & \textbf{0.65{\tiny$\pm$.07}} & \textbf{0.67{\tiny$\pm$.07}} & 0.63{\tiny$\pm$.07} & \textbf{0.64{\tiny$\pm$.06}} & \textbf{0.64{\tiny$\pm$.06}} \\
        \bottomrule
    \end{tabular}
    \caption{Simulation accuracy across persona pools (mean $\pm$ SE). Full version of Table~\ref{tab:results:personas}.}
    \label{tab:app:personas_full}
\end{table*}

\begin{table*}[h]
    \centering
    \small
    \setlength{\tabcolsep}{3pt}
    \begin{tabular}{l ccc ccc ccc}
        \toprule
        & \multicolumn{3}{c}{\textbf{CTR} ($n{=}20$)} & \multicolumn{3}{c}{\textbf{Subscriptions} ($n{=}20$)} & \multicolumn{3}{c}{\textbf{Combined} ($n{=}40$)} \\
        \cmidrule(lr){2-4} \cmidrule(lr){5-7} \cmidrule(lr){8-10}
        \textbf{Algorithm} & Acc & SignOv & SignBC & Acc & SignOv & SignBC & Acc & SignOv & SignBC \\
        \midrule
        Random          & 0.79{\tiny$\pm$.003} & 0.74{\tiny$\pm$.003} & 0.75{\tiny$\pm$.003} & 0.91{\tiny$\pm$.002} & 0.78{\tiny$\pm$.003} & 0.79{\tiny$\pm$.003} & 0.85{\tiny$\pm$.002} & 0.76{\tiny$\pm$.002} & 0.77{\tiny$\pm$.002} \\
        Kernel Herding  & 0.78{\tiny$\pm$.001} & 0.74{\tiny$\pm$.001} & 0.75{\tiny$\pm$.001} & \textbf{0.91{\tiny$\pm$.001}} & \textbf{0.78{\tiny$\pm$.001}} & \textbf{0.80{\tiny$\pm$.001}} & 0.84{\tiny$\pm$.001} & 0.76{\tiny$\pm$.001} & \textbf{0.78{\tiny$\pm$.001}} \\
        Greedy Farthest & \textbf{0.80{\tiny$\pm$.002}} & \textbf{0.74{\tiny$\pm$.002}} & \textbf{0.75{\tiny$\pm$.002}} & 0.90{\tiny$\pm$.001} & 0.77{\tiny$\pm$.002} & 0.80{\tiny$\pm$.002} & \textbf{0.85{\tiny$\pm$.001}} & \textbf{0.76{\tiny$\pm$.001}} & \textbf{0.78{\tiny$\pm$.001}} \\
        \midrule
        All (935)       & 0.80 & 0.74 & 0.75 & 0.90 & 0.80 & 0.81 & 0.85 & 0.77 & 0.78 \\
        \bottomrule
    \end{tabular}
    \caption{Subsampling performance at 500 personas (mean $\pm$ SE across 1000 trials). Full version of Table~\ref{tab:results:subsampling}.}
    \label{tab:app:subsampling_full}
\end{table*}

\begin{table*}[h]
    \centering
    \small
    \setlength{\tabcolsep}{3pt}
    \begin{tabular}{l ccc ccc ccc}
        \toprule
        \multirow{2}{*}{\textbf{Configuration}} & \multicolumn{3}{c}{\textbf{CTR} ($n{=}20$)} & \multicolumn{3}{c}{\textbf{Subscriptions} ($n{=}20$)} & \multicolumn{3}{c}{\textbf{Combined} ($n{=}40$)} \\
        \cmidrule(lr){2-4} \cmidrule(lr){5-7} \cmidrule(lr){8-10}
        & Acc & SignBC & SignOv & Acc & SignBC & SignOv & Acc & SignBC & SignOv \\
        \hline
        Pairwise rating \emph{(full)}   & \textbf{0.70{\tiny$\pm$.11}} & \textbf{0.64{\tiny$\pm$.09}} & \textbf{0.64{\tiny$\pm$.09}} & \textbf{0.80{\tiny$\pm$.09}} & 0.72{\tiny$\pm$.07} & 0.71{\tiny$\pm$.07} & \textbf{0.75{\tiny$\pm$.07}} & \textbf{0.68{\tiny$\pm$.06}} & \textbf{0.68{\tiny$\pm$.06}} \\
        Demographics only \emph{(full)} & 0.65{\tiny$\pm$.11} & 0.63{\tiny$\pm$.09} & 0.57{\tiny$\pm$.08} & 0.30{\tiny$\pm$.11} & 0.51{\tiny$\pm$.07} & 0.39{\tiny$\pm$.07} & 0.47{\tiny$\pm$.08} & 0.57{\tiny$\pm$.06} & 0.48{\tiny$\pm$.05} \\
        No reasoning \emph{(full)}      & 0.60{\tiny$\pm$.11} & 0.61{\tiny$\pm$.09} & 0.60{\tiny$\pm$.09} & \textbf{0.80{\tiny$\pm$.09}} & \textbf{0.76{\tiny$\pm$.06}} & \textbf{0.73{\tiny$\pm$.06}} & 0.70{\tiny$\pm$.07} & 0.69{\tiny$\pm$.05} & 0.67{\tiny$\pm$.05} \\
        Generic shopper \emph{(single)} & 0.40{\tiny$\pm$.11} & 0.44{\tiny$\pm$.10} & 0.44{\tiny$\pm$.10} & 0.60{\tiny$\pm$.11} & 0.65{\tiny$\pm$.08} & 0.64{\tiny$\pm$.08} & 0.50{\tiny$\pm$.08} & 0.55{\tiny$\pm$.06} & 0.54{\tiny$\pm$.06} \\
        No persona \emph{(single)}      & 0.45{\tiny$\pm$.11} & 0.49{\tiny$\pm$.10} & 0.49{\tiny$\pm$.10} & 0.65{\tiny$\pm$.11} & 0.65{\tiny$\pm$.08} & 0.63{\tiny$\pm$.08} & 0.55{\tiny$\pm$.08} & 0.57{\tiny$\pm$.06} & 0.56{\tiny$\pm$.06} \\
        \bottomrule
    \end{tabular}
    \caption{\wl{Component contributions to simulation accuracy (mean $\pm$ SE). Full version of Table~\ref{tab:results:ablation_baselines}. Full pool baselines use 935 personas \emph{(full)}; single-call baselines use one fixed prompt \emph{(single)}.}}
    \label{tab:app:ablation_full}
\end{table*}

\section{Comparison with SimAB Benchmark}
\label{app:simab}

We investigated the feasibility of benchmarking our framework against SimAB~\cite{Rieder:2026:arXiv_SimAB}.
SimAB reports 67\% overall accuracy (83\% for high-confidence cases) on a 47-test corpus,
of which the Wikimedia Foundation fundraising tests (2010--2011) constitute the only publicly
reproducible subset. These tests involve banner designs and landing page variants for Wikipedia
donation campaigns, with ground truth derived from documented conversion metrics
(donations/impression, amount/impression, click-through rate).

\paragraph{Reproducibility barriers.}
SimAB sources its experimental data from the Wikimedia 2010 Banner
Testing\footnote{\url{https://meta.wikimedia.org/wiki/Fundraising_2010/Banner_testing}} and
2011 Test Updates\footnote{\url{https://meta.wikimedia.org/wiki/Fundraising_2011/Test_Updates}}
pages. However, the authors do not specify which subset of tests from these sources were used,
nor do they publish a code repository or processed datasets. As a result, there is no reliable
way to identify the exact experimental conditions under which their reported numbers were
obtained, precluding a direct numerical comparison.
Beyond reproducibility, several fundamental methodological differences complicate a fair
comparison, summarised in Table~\ref{tab:simab_comparison}.

\begin{table*}[t]
    \centering
    \small
    \begin{tabular}{lll}
        \toprule
        \textbf{Dimension} & \textbf{Our Framework} & \textbf{SimAB} \\
        \midrule
        Persona source & Data-driven (behavioural) & Synthetic (from screenshots) \\
        Question format & Pairwise rating (1--10) & Binary preference (C/Ch/None) \\
        Bias mitigation & Random order per persona & Deterministic counterbalancing \\
        Population size & Fixed (100 personas) & Variable (30--200, early stopping) \\
        Aggregation & Mean effect + SE $\to$ Normal dist. & Majority vote $\to$ significance test \\
        Primary metric & Directional accuracy + SignBC & Accuracy + F1 \\
        Stimuli & \wl{Visual screenshots + textual question framing} & Visual screenshots \\
        Ground truth source & Experimental ground-truth labels & Documented campaign outcomes \\
        \bottomrule
    \end{tabular}
    \caption{Key methodological differences between our framework and SimAB.}
    \label{tab:simab_comparison}
\end{table*}

\section{Population Subsampling: Extended Results}
\label{app:subsampling}

We compute dense vector embeddings for each persona using Qwen2.5-32B-Instruct \citep{qwen2.5}, which captures semantic characteristics of persona profiles. The subsampling algorithms then operate on these embeddings:
\begin{itemize}[leftmargin=1em, itemsep=0pt, parsep=0pt, topsep=3pt]
\item \textbf{Random}: Uniform random sampling without replacement
\item \textbf{Greedy Farthest}: Iteratively selects personas maximizing minimum distance to already selected personas, ensuring geometric diversity by spreading selected points across the embedding space. We randomize the first point to enable variation across trials; subsequent selections are deterministic given the first point.
\item \textbf{Kernel Herding}: Sequentially selects personas to minimize Maximum Mean Discrepancy (MMD) between the subsample and the full distribution \citep{chen2012super}. While kernel herding is inherently deterministic, we implement a variant that randomizes the first point to generate diverse trials; all subsequent selections are deterministic.
\end{itemize}

\begin{figure}[h]
    \centering
    \includegraphics[width=\linewidth]{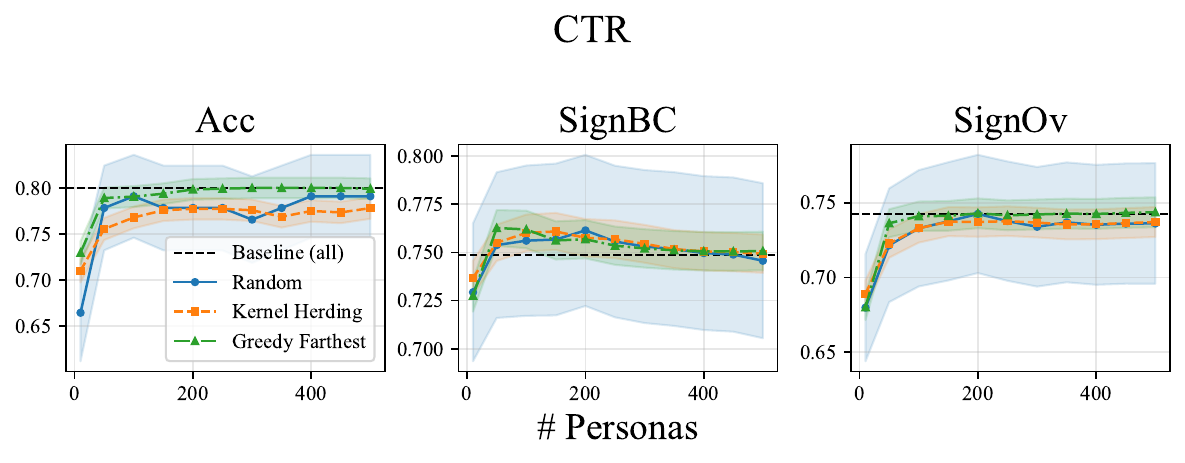}
    \caption{Marketing alignment metrics vs sample size for the different sampling algorithms.}
    \label{fig:marketing-sampling}
\end{figure}

\begin{figure}[h]
    \centering
    \includegraphics[width=\linewidth]{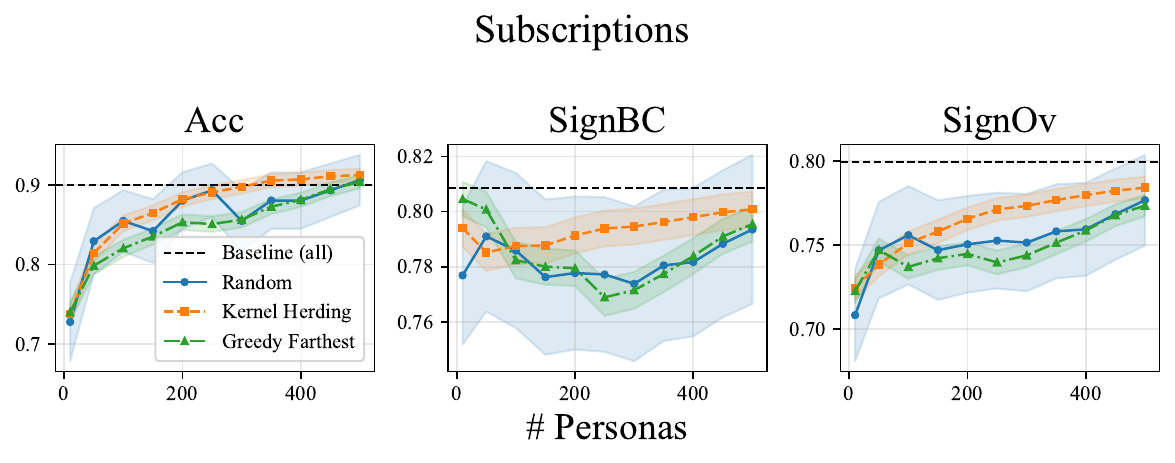}
    \caption{Subscription alignment metrics vs sample size for the different sampling algorithms.}
    \label{fig:subscriptions-sampling}
\end{figure}

Figures \ref{fig:marketing-sampling} and \ref{fig:subscriptions-sampling} show the evolution of the sign metrics with the size of the sample, comparing the performance of the three algorithms. Each point in the figures corresponds to the average over 1000 independent trials, with error bars representing standard error (SE). For CTR tests, Figure \ref{fig:marketing-sampling} shows that even a small sample of size 100 can recover the performance of the full pool of personas, with Farthest Greedy slightly outperforming Random sampling and Kernel Herding. For subscription tests, Figure \ref{fig:subscriptions-sampling} suggests that Kernel-herding has the best performance among the three algorithms, but the performance gap is again small.

For both use-cases, even random sampling gives strong results. A main advantage of Greedy Farthest or Kernel Herding is their small variance compared to random sampling.

Note that in some cases, the sampled personas outperform the baseline of using the entire pool. This is because the full persona set may produce overly confident predictions that, while directionally correct, exhibit excessive certainty compared to the true distribution's inherent uncertainty. This overconfidence can harm uncertainty-aware metrics such as SignOv and SignBC. Subsampling introduces natural variance that can yield closer distributions to the ground truth with appropriate uncertainty levels. Furthermore, when the full set's prediction is directionally incorrect, the added uncertainty from fewer personas can actually improve alignment with ground truth.

\section{Implementation Details}
\label{app:implementation}

\noindent
The prompts and templates in the sections below are slightly simplified for clarity; the final versions used in the experiments include additional formatting instructions and edge-case handling.

\subsection{Model Configuration}
  
All experiments use Claude Sonnet 4.5~\citep{Anthropic:2025:Sonnet4.5} accessed via cloud-based batch inference with the following parameters:
\begin{itemize}[leftmargin=1em, itemsep=0pt, parsep=0pt, topsep=3pt]
    \item Temperature: 0 (for reproducibility)
    \item Max tokens: 3,000
    \item Response format: JSON (enforced via assistant prefill \texttt{<JSON> \{})
\end{itemize}

\subsection{Persona Generation}

Personas are generated through two sequential LLM flows:
\begin{enumerate}[leftmargin=1em, itemsep=0pt, parsep=0pt, topsep=3pt]
    \item \textbf{Demographics flow}: Infers age, gender, income, education, household composition, and other attributes from behavioral patterns using structured output schemas with constrained category options.
    \item \textbf{Shopping characteristics flow}: Synthesizes narrative descriptions of brand affinity, price sensitivity, quality expectations, and shopping style from the same behavioral signals.
\end{enumerate}

\noindent
Both flows use Claude Sonnet 4.5 with structured JSON output schemas to ensure consistent persona formats across the pool.

\subsection{Persona Template}
\label{app:implementation_personas}

We construct each persona by processing anonymized, aggregated behavioral patterns through an LLM, which synthesizes these signals into a structured profile.\footnote{These metric and tier names are generic e-commerce terminology chosen by the authors for readability.}
The profile captures multiple complementary dimensions:
\begin{itemize}[leftmargin=1em, itemsep=0pt, parsep=0pt, topsep=3pt]
    \item \textbf{Inferred demographics}: attributes such as age, gender, income, and household composition, derived from behavioral patterns rather than self-reported.
    \item \textbf{Behavioral statistics}: quantitative summaries of activity intensity, spending patterns, and category preferences.
    \item \textbf{Engagement patterns}: conversion funnels and temporal regularities that characterize how the user interacts with the platform.
    \item \textbf{LLM-synthesized narratives}: free-text interest profiles and behavioral insight summaries that provide the model with a coherent characterization to role-play during simulation.
\end{itemize}
\noindent
During simulation, the full persona profile is rendered into a structured prompt that instructs the LLM to embody the corresponding user---maintaining first-person perspective, using language appropriate to their background, and making decisions consistent with their established behavior patterns. Fields are populated from the persona generation pipeline. 

\begin{small}
\begin{verbatim}
## ROLE INSTRUCTION
You are now embodying the persona of a specific
online retail customer. Respond as if you were
this person, with their demographic background,
interests, and behavioral patterns. Maintain
first-person perspective throughout.

## CUSTOMER PROFILE

### DEMOGRAPHIC BACKGROUND
- Age: {age}
- Gender: {gender}
- Education: {education}
- Income Level: {income}
- Home Status: {home_ownership}
- Family: {parental_status}
- Location: {urbanicity}

### INTERESTS AND PREFERENCES
{interests}

### SHOPPING BEHAVIOR PROFILE
- Brand Perspective: {brand_perception}
- Price Sensitivity: {price_perception}
- Quality Expectations: {quality_perception}
- Use of Reviews: {reviews_perception}
- Value Assessment: {value_perception}

### PURCHASE HISTORY (aggregated)
- Activity Level: {activity_tier}
- Purchase Frequency: {frequency_tier}
- Spending Tier: {spending_tier}
- Price Range: {price_range}

### ENGAGEMENT METRICS
- Search to View Rate: {search_to_view}
- View to Purchase Rate: {view_to_purchase}

## BEHAVIORAL INSIGHTS
{behavioral_reasoning}

## EMBODIMENT GUIDELINES
1. Use language appropriate for your background.
2. Incorporate your characteristic approach to
 evaluating products.
3. Express opinions consistent with your profile.
\end{verbatim}
\end{small}
  
\subsection{Question Prompt Template (Pairwise Rating)}
  
\begin{small}
\begin{verbatim}
System: You are an AI assistant role-playing a
customer with specific characteristics. You are
answering a structured question about visual
design variants.

User:
1. Background about your character:
<background>{persona}</background>

2. Your task: Rate each option on a 1-10 scale.
Provide your scores in the "answer" field and
a reason in the "reason" field.

3. Response format (valid JSON):
{"reason": "explanation grounded in persona",
"answer": "widget_1:7, widget_2:4"}

<question>{question_text}</question>
<options>
1: {variant_label_1} [image]
2: {variant_label_2} [image]
</options>
\end{verbatim}
\end{small}

\noindent
The agent receives a system prompt instructing it to role-play as the persona, followed by the structured question with variant screenshots. Images are encoded as base64 images and passed inline within the prompt as image content blocks, allowing the model to process visual stimuli directly alongside the textual question and persona context.
For independent rating, each variant is presented in a separate prompt without alternatives. Tested variant order is randomized per persona using the persona identifier as a deterministic seed, ensuring reproducibility while controlling for positional bias. Variant labels use neutral identifiers (\texttt{widget\_\{n/m\}}) to avoid priming effects. Each simulation run processes all personas in parallel via batch jobs, with automatic batching based on estimated prompt size. 

\subsection{Score Extraction}

Scores are extracted from the JSON response's \texttt{answer} field by parsing the \texttt{widget\_name:score} pairs. For each A/B test, we collect per-persona $(s_{ref}, s_{treat})$ tuples and compute the predicted effect as $\hat{\delta}_s = \frac{1}{N}\sum_{i=1}^{N} \frac{s_{treat}^{(i)} - s_{ref}^{(i)}}{s_{ref}^{(i)}}$.

\end{document}